%% file: main.tex
\documentclass[11pt]{article}
\usepackage[letterpaper,margin=0.86in]{geometry}
\usepackage[T1]{fontenc}
\usepackage[utf8]{inputenc}
\usepackage{newtxtext,newtxmath}
\usepackage{microtype,graphicx,booktabs,tabularx,array,enumitem}
\usepackage{amsmath,mathtools}
\usepackage[dvipsnames]{xcolor}
\usepackage[numbers,sort&compress]{natbib}
\usepackage{tikz}
\usetikzlibrary{arrows.meta,positioning,fit,calc}
\usepackage{caption}
\usepackage{placeins}
\usepackage{hyperref}
\hypersetup{colorlinks=true,linkcolor=MidnightBlue,citecolor=MidnightBlue,urlcolor=MidnightBlue,
pdftitle={MSR: Multiple Subject Reference for Video Generation},pdfauthor={Guannan Li, Jiaji Chen, Jingyuan Liao, Yu Geng, Baolan Qiu}}
\usepackage[nameinlink,noabbrev]{cleveref}
\setlist[itemize]{leftmargin=*,itemsep=2pt,topsep=4pt}
\newcommand{\R}{\mathbb{R}}
\newcommand{\code}[1]{\texttt{\small #1}}
\title{\vspace{-1.5em}\textbf{MSR: Multiple Subject Reference\\ for Video Generation}}
\author{Guannan Li\quad Jiaji Chen\\
Jingyuan Liao\thanks{Jingyuan Liao, Yu Geng, and Baolan Qiu contributed equally.}\quad Yu Geng\footnotemark[1]\quad Baolan Qiu\footnotemark[1]\\
Licon Studio\\\href{mailto:lign2@hotmail.com}{\texttt{lign2@hotmail.com}}}
\date{September 2026}
\begin{document}
\maketitle
\begin{abstract}
Conditioning a video generator on multiple images requires preserving appearance while associating each reference with its intended role. We present MSR (Multiple Subject Reference), a slot-aware conditioning scheme for LTX-based video generation. Each reference image is independently encoded as a static clip and represented by a separate latent-token group. A compact Fourier-feature multilayer perceptron adds a numeric slot embedding, while slot-dependent temporal offsets modify the group's rotary coordinates. The reference groups are prepended to noisy target tokens and serve as clean context during target-only flow-matching training. We implement this scheme through low-rank adaptation and release the resulting weights and inference workflows. Qualitative examples demonstrate compositions containing distinct characters and referenced environments in realistic and stylized scenes. Development observations suggest reduced reference confusion relative to an earlier continuous-reference baseline, while similar clothing, complex garments, and viewpoint changes remain challenging. We describe the conditioning mechanism, the retained training configuration, and the observed strengths and limitations of the released system. A supplementary audio-reference experiment adds voice conditioning while keeping the visual parameters frozen.
\end{abstract}
\begin{center}\small
\href{https://huggingface.co/LiconStudio/LTX-2.5-Multiple-Subject-Reference}{Model and workflows}\quad$\cdot$\quad
\href{https://huggingface.co/spaces/hugging-apps/ltx25-multi-subject-reference}{Interactive demonstration}
\end{center}
\input{sections/introduction}
\input{sections/method}
\input{sections/experiments}
\input{sections/discussion}
{\footnotesize
\bibliographystyle{plainnat}
\bibliography{references}
}
\clearpage
\appendix
\small
\input{sections/appendix}
\end{document}

%% file: sections/introduction.tex
\section{Introduction}
Reference-conditioned video generation is useful when a scene must contain specified visual elements rather than arbitrary instances of a text description. In a two-character interaction, for example, a prompt may need to preserve each character's face, hairstyle, and clothing, place both in a referenced environment, and maintain the assignment of attributes as the camera changes viewpoint. Producing a plausible frame is insufficient if one character acquires the other's jacket or the referenced object changes ownership.

The problem is not limited to facial identity. A reference may specify a complete character, a garment, a prop, or a setting. Several views can also be arranged within a single reference image. Such an image still constitutes one input source; its constituent views are not automatically separate subjects. A useful interface should preserve the association between an input image and the role assigned to it in the prompt, while allowing the generator to create new poses, compositions, and interactions.

An early version of our system assembled reference images into a continuous pseudo-video and concatenated the resulting reference representation with the generation context. This provides a simple route for reusing a video model, but it does not explicitly identify independent reference sources. In particular, adjacent frames of the pseudo-video can depict unrelated subjects even though a video encoder is designed to process temporally related content. This observation motivates separating image encoding from the organization of the reference context.

Multiple Subject Reference (MSR) represents each input image as an independently encoded token group. It supplies two complementary source cues: a learned numeric slot embedding in latent-token space, and a deterministic shift of the group's temporal positional coordinates. The former changes the reference features; the latter changes their positional relationships inside the existing transformer. Ordinary text conditioning specifies the roles and actions associated with the input images. We adapt the LTX model family~\citep{LTX22026} using LoRA~\citep{LoRA2021} and a small slot module, without adding a separate face-recognition or multimodal-language-model conditioning branch.

This paper makes three contributions. First, it specifies an implementable slot-aware conditioning scheme, including the exact slot representation, token organization, and positional convention. Second, it documents the corresponding visual training configuration and provides publicly accessible weights and workflows. Third, it analyzes qualitative behavior across multi-reference cases, development comparisons, and observed failure modes.

The main study concerns visual multi-subject conditioning. A subsequent audio-reference extension freezes the visual parameters and trains audio-related modules. We describe it in the appendix to separate the visual method from additional audio capabilities.

\section{Related Work}
\paragraph{Video foundations and efficient adaptation.}
LTX-2 provides a joint audio--video generation architecture with interacting visual and audio streams~\citep{LTX22026}. Our implementation adapts a pretrained backbone from the LTX model family. LoRA introduces low-rank updates to pretrained weight matrices~\citep{LoRA2021}, and flow matching provides the general velocity-regression formulation used for generation~\citep{FlowMatching2022}. We retain these foundations and focus on representing multiple reference sources within the visual conditioning sequence.

\paragraph{Subject-conditioned video generation.}
Ingredients combines facial extraction, projection, and identity routing to condition video diffusion transformers on multiple human identities~\citep{Ingredients2025}. MAGREF addresses heterogeneous references through composite reference layouts, masked guidance, and channel-wise conditioning; its subject-disentanglement formulation also associates semantic subject information with visual regions~\citep{MAGREF2025}. BindWeave uses multimodal language-model hidden states to connect reference subjects with prompt semantics before conditioning a video generator~\citep{BindWeave2025}. MSR instead exposes independently encoded visual token groups to the existing transformer and retains its text-conditioning pathway.

\paragraph{Reference identity and positional conditioning.}
Rotary position embeddings introduce position-dependent relationships in attention~\citep{RoFormer2021}. As a separate related method, ID-LoRA uses negative reference temporal positions for LTX-based audiovisual personalization~\citep{IDLoRA2026}. Aura combines reference-token concatenation, learned category information, rotary-coordinate shifts, and a VLM-based conditioning pathway~\citep{Aura2026}. MSR shares the motivation of distinguishing reference sources, but uses a numeric image-slot MLP and a selected-slot temporal offset within an LTX adaptation. Our contribution is a concrete slot-indexed realization of these conditioning principles for parameter-efficient LTX adaptation.

%% file: sections/method.tex
\section{Method}
\label{sec:method}

\subsection{Task and reference representation}
Let $p$ be a text prompt and $\{I_{s_i}\}_{i=1}^{K}$ a set of supplied images ordered by their slot identifiers $s_i$. The task is to generate a video whose content follows $p$ and preserves the relevant appearance and role of each reference. A slot is an input-source identifier, not a semantic class or a guarantee of one entity per image. The interface supports human and stylized characters as well as clothing, objects, and scenes, with the intended use specified in text.

For each image, we form a static clip by repeating only that image for $F_r$ frames and apply the backbone's video VAE encoder $E$ independently:
\begin{equation}
 R_i=P\!\left(E\!\left(\operatorname{repeat}(I_{s_i},F_r)\right)\right)
 \in\R^{N_i\times d_r},
 \label{eq:encode}
\end{equation}
where $P$ denotes latent patchification and $N_i$ is the number of tokens for that reference. Different images never share an encoder input clip. A multi-view collage already contained in one image remains a single reference. The archived training preprocessing uses $F_r=25$ at 25 frames per second. The reference token dimension for the released slot module is $d_r=128$.

Independent encoding separates the VAE inputs of unrelated references. It does not prevent interactions between them after token concatenation, and it does not impose an attention mask. That downstream interaction is necessary for composing a scene from several input sources.

\subsection{Numeric slot embeddings}
Each reference group receives an embedding determined by its positive integer image-slot identifier. For $u_s=s/16$ and $\omega_j=2^{4j/15}$, $j\in\{0,\ldots,15\}$, we define
\begin{align}
 \phi(s)&=\big[u_s,\{\sin(u_s\omega_j)\}_{j=0}^{15},
                  \{\cos(u_s\omega_j)\}_{j=0}^{15}\big]\in\R^{33},\\
 e_s&=W_2\operatorname{SiLU}(W_1\phi(s)+b_1)+b_2\in\R^{128},
 \label{eq:slot}\\
 \widehat R_i&=R_i+\mathbf{1}_{N_i}e_{s_i}^{\mathsf T}.
\end{align}
The MLP has dimensions $33\!\rightarrow\!256\!\rightarrow\!128$, with biases in both layers and no normalization layer. It contains 41,600 learned parameters; its 16 frequency values are fixed buffers. The embedding is added before the transformer's input projection, so every token in a reference group carries the same source tag without changing its sequence length. Target tokens do not receive a reference-slot embedding.

The module encodes an index rather than a learned category such as ``person'' or ``background.'' The text prompt supplies these roles. Although the MLP can be evaluated for integer IDs beyond those encountered in training, this algebraic property is not evidence of reliable generation with an unlimited number of references.

\subsection{Slot-dependent temporal offsets}
\label{sec:offset}
The video transformer also receives spatial--temporal coordinates used by its positional encoding. Let $\bar\tau_{i,n}$ be a temporal coordinate for reference token $n$ after the implementation's reference-to-target alignment, and let $f_v$ denote the target frame rate. For $K$ selected references in natural slot order, we use
\begin{equation}
 \Delta_i=-\frac{K-i+1}{f_v},\qquad
 \tau'_{i,n}=\bar\tau_{i,n}+\Delta_i,
 \quad i=1,\ldots,K.
 \label{eq:offset}
\end{equation}
The shift is applied to both temporal endpoints when a token is represented by a coordinate interval. Spatial coordinates retain the reference-to-target scale adjustment. The embedding in \cref{eq:slot} uses the original slot ID $s_i$, whereas \cref{eq:offset} uses the compact order $i$ among selected references. They agree for consecutively numbered inputs.

These are negative \emph{offsets}, not disjoint negative temporal windows. For three references at 25 fps, the offsets are $-0.12$, $-0.08$, and $-0.04$ seconds; a reference with nonzero temporal extent can still have positive coordinates after translation. The shift distinguishes positional context but does not constrain an attention matrix or guarantee identity separation. Positional coordinates also differ from diffusion timesteps: the former indicate where a token lies in the model's coordinate system, while the latter indicate its noise level.

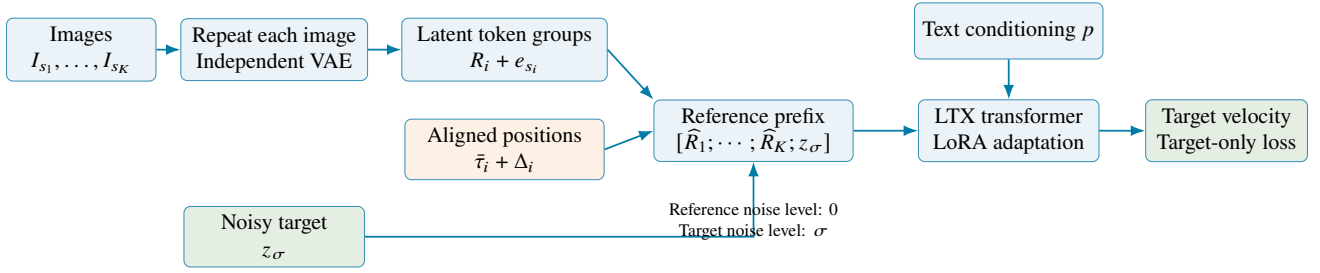
\begin{figure}[t]
\centering
\resizebox{\linewidth}{!}{%
\begin{tikzpicture}[font=\small,>=Latex,
 box/.style={draw=MidnightBlue!65,rounded corners=3pt,fill=MidnightBlue!5,align=center,minimum height=9mm},
 line/.style={->,thick,draw=MidnightBlue!80}]
\node[box,minimum width=24mm] (im) at (0,1.3) {Images\\$I_{s_1},\ldots,I_{s_K}$};
\node[box,minimum width=30mm] (enc) at (3.1,1.3) {Repeat each image\\Independent VAE};
\node[box,minimum width=33mm] (tag) at (6.8,1.3) {Latent token groups\\$R_i+e_{s_i}$};
\node[box,minimum width=32mm,fill=Orange!10] (pos) at (6.8,-.3) {Aligned positions\\$\bar\tau_i+\Delta_i$};
\node[box,minimum width=29mm,fill=OliveGreen!10] (noisy) at (3.1,-1.7) {Noisy target\\$z_\sigma$};
\node[box,minimum width=32mm] (seq) at (10.8,.0) {Reference prefix\\$[\widehat R_1;\cdots;\widehat R_K;z_\sigma]$};
\node[box,minimum width=29mm] (dit) at (14.9,.0) {LTX transformer\\LoRA adaptation};
\node[box,minimum width=26mm,fill=OliveGreen!10] (out) at (18.4,.0) {Target velocity\\Target-only loss};
\node[box,minimum width=29mm] (text) at (14.9,1.6) {Text conditioning $p$};
\draw[line] (im)--(enc);\draw[line] (enc)--(tag);
\draw[line] (tag.east)--(seq.north west);
\draw[line] (pos.east)--(seq.west);
\draw[line] (noisy.east)-|(seq.south);
\draw[line] (seq)--(dit);\draw[line] (text)--(dit);\draw[line] (dit)--(out);
\node[align=center,font=\scriptsize] at (10.8,-1.45) {Reference noise level: $0$\\Target noise level: $\sigma$};
\end{tikzpicture}}
\caption{MSR visual conditioning. Each reference is encoded separately, tagged in latent-token space, and assigned a slot-dependent positional offset. Clean reference tokens precede noisy target tokens. Only the target suffix contributes to the training loss. The diagram abstracts the existing transformer and its text-conditioning path.}
\label{fig:method}
\end{figure}

\subsection{Clean context and target-only training}
Given target video tokens $z_0\in\R^{N_v\times d_r}$ and Gaussian noise $\epsilon$, we sample a noise level $\sigma$ and interpolate
\begin{equation}
 z_\sigma=(1-\sigma)z_0+\sigma\epsilon,\qquad
 v^\star=\epsilon-z_0.
\end{equation}
The transformer receives the concatenated sequence
\begin{equation}
 X_\sigma=[\widehat R_1;\ldots;\widehat R_K;z_\sigma].
 \label{eq:concat}
\end{equation}
Reference tokens remain clean and are assigned diffusion timestep zero; target tokens receive timestep $\sigma$. Conditioning includes the text representation and the spatial--temporal coordinates described above. We retain the predicted target suffix and minimize the velocity-regression loss
\begin{equation}
 \mathcal L=\mathbb E_{z_0,\epsilon,\sigma}
 \left[\frac{1}{N_vd_r}
 \left\|v_\theta(X_\sigma,p)_{\rm target}-(\epsilon-z_0)\right\|_F^2\right].
 \label{eq:loss}
\end{equation}
This is the all-target-token case of the implementation's normalized masked loss, following the flow-matching formulation~\citep{FlowMatching2022}. Reference tokens provide conditioning but no reconstruction-loss terms.

Training updates LoRA adapters on the visual self-attention and text cross-attention query, key, value, and output projections, as well as the visual feed-forward input and output projections. The reference-slot MLP is trainable. The underlying backbone weights remain frozen. At inference, the reference context and slot correspondence are constructed before denoising; a normal text prompt identifies the input images and the desired scene.

\subsection{Relationship to the earlier baseline}
\Cref{tab:versions} distinguishes the earlier continuous-reference representation from the released visual MSR. The changes form a combined method comparison: independent image encoding, slot embeddings, and positional offsets change together. Consequently, a favorable version-level observation cannot isolate the gain from any single component.
\begin{table}[t]
\centering\small
\begin{tabularx}{\linewidth}{@{}lXX@{}}\toprule
Property & Earlier MSR V1/V2 & Released visual MSR\\\midrule
Training backbone & LTX & LTX\\
Reference encoding & Continuous pseudo-video of input images & Independent static clip per image\\
Learned image-slot MLP & Absent & Present\\
Slot-dependent negative offset & Absent & Present\\
\bottomrule
\end{tabularx}
\caption{Version definitions. Both variants use the same LTX training backbone. Comparisons in development used the same inference backbone; this table describes the combined version changes, not isolated ablations.}
\label{tab:versions}
\end{table}

%% file: sections/experiments.tex
\section{Experiments and Qualitative Analysis}
\label{sec:experiments}
\subsection{Training data and implementation}
The documented visual training stage uses examples with English captions, a target video, and numbered reference images. The training mixture comprises approximately 28.91\% from \code{MSR\_batch2\_V2}, 36.61\% from \code{MSR\_training\_data\_V6}, and 34.47\% from \code{short\_drama\_V2}; \code{MSR\_batch4\_V1} is excluded from this stage. Each caption describes the supplied sources and the requested scene or actions. The records cover different reference counts and visual roles. \Cref{tab:refs} reports the percentage of examples at each nonempty reference-image slot count in the metadata entries matched to this manifest.

\begin{table}[t]
\centering\small
\begin{tabular}{@{}lrrrrrrr@{}}\toprule
Reference images per example & 2 & 3 & 4 & 5 & 6 & 7 & 8\\\midrule
Share of examples (\%) & 20.35 & 25.99 & 29.53 & 23.89 & 0.12 & 0.08 & 0.04\\\bottomrule
\end{tabular}
\caption{Distribution of reference-image counts in the visual-stage manifest, expressed as percentages of examples. Reference counts refer to image slots, not the number of subjects or views inside an image. Percentages are rounded to two decimal places and may not sum to 100\%. Unique file paths do not establish content-level deduplication.}
\label{tab:refs}
\end{table}

Reference preprocessing repeats an image for 25 frames at 25 fps and independently encodes it using the video VAE. The archived scripts provide landscape, portrait, and square resolution buckets. Target videos are processed at their source frame rates and trimmed to a supported $8n+1$ frame count without temporal padding. The documented preprocessing job and metadata schema use up to eight reference-image fields, while the released visual interface exposes up to five input images. The few training examples with six to eight images do not constitute an evaluation of generation quality at those counts.

\Cref{tab:train} summarizes the retained configuration for the visual stage. The public visual adapter is an intermediate checkpoint from that stage. Its exact step number was not retained locally, so neither the later saved 13,000-step checkpoint nor the 20,000-step configuration ceiling is used as the release's training duration. The retained configuration includes a subsequent resume; the table documents that stage rather than reconstructing an exact schedule for the released intermediate file.

\begin{table}[t]
\centering\small
\begin{tabularx}{\linewidth}{@{}lX@{}}\toprule
Setting & Retained visual-stage configuration\\\midrule
Backbone & LTX (22B)\\
Text encoder & Gemma 3 12B IT\\
LoRA rank / scaling $\alpha$ / dropout & 128 / 128 / 0\\
Slot MLP & $33\rightarrow256\rightarrow128$, SiLU, 41,600 parameters\\
Optimizer / learning rate & AdamW / $10^{-4}$\\
Schedule / precision & Cosine / BF16\\
Parallelism / batch & 8 devices; 1 example per device; accumulation 1\\
Gradient clipping / checkpointing & Norm 1.0 / enabled\\
Sampling & No bucket-aware sampling\\
Noise sampling & Two-region shifted-logit-normal mixture (Appendix~\ref{app:settings})\\
Configured seed & 42\\\bottomrule
\end{tabularx}
\caption{Settings recorded for the stage that produced the released visual adapter. The exact public checkpoint step and a full historical schedule are not reconstructed from a later resume configuration.}
\label{tab:train}
\end{table}

\subsection{Evaluation scope}
Our evaluation is qualitative. It consists of public reference--video pairs and observations made during method development. We show cases explicitly associated with the released visual adapter in its model card, rather than infer model identities from unrelated local filenames. For each displayed public clip, we sample three frames at 15\%, 50\%, and 85\% of its 12.04-second duration. Selection is illustrative, not random, and three frames cannot establish consistency at every instant of a video.

The earlier and newer adapters were compared on the same inference backbone during development. However, a complete archive of paired runs, random seeds, and per-trial judgments was not retained. We therefore summarize these observations without a numerical success rate, statistical significance claim, or quantitative ranking against other methods. An additional local demonstration collection contains selected complete cases, with multiple outputs for some inputs; it is not an exhaustive test set. We do not use that collection to infer a population success rate.

\subsection{Composing distinct character and scene references}
\Cref{fig:realistic} shows a public example with two separate character references and a nightclub reference. Across the sampled frames, the dark jacket of one character and the silver outfit and transparent outer garment of the other remain distinguishable, while the composition changes from a wider view to closer interaction. The nightclub lighting and booth environment are also reflected in the generated frames. This illustrates the intended task: several referenced elements participate in a newly composed interaction rather than appearing as independent still images.

\begin{figure}[p]
\centering
\input{figures/case_03.tex}
\caption{Public visual MSR case 03. Top: two character references and a scene reference. Bottom: full generated frames sampled at approximately 1.83, 6.04, and 10.25 seconds. The images show distinguishable appearance cues and the referenced setting across changes in framing. Frames are not cropped; the composite views in each character reference were already present in the input image. }
\label{fig:realistic}
\vspace{1em}
\input{figures/case_06.tex}
\caption{Public visual MSR case 06. Top: two stylized-character references and a forest reference. Bottom: the same relative sampling times as in \cref{fig:realistic}. The sampled views show the contrast between the orange-haired character in a light dress and the silver-haired character in a green outfit while changing camera distance. }
\label{fig:stylized}
\end{figure}

\Cref{fig:stylized} uses a different visual style: two stylized characters and a forest setting. The sampled output includes a wide environment view and closer two-character views. Hair color, garment color, and distinctive head features provide visible cues for separating the two references. Together, the two cases demonstrate that the interface can condition on more than facial portraits: each character image includes overall appearance, and the third image supplies an environment.

Scene layout, pose, perspective, and illumination are recomposed by the generator. The figures illustrate reference-conditioned composition rather than exact replication of the input images.

\subsection{Development observations and failure modes}
In development comparisons against the earlier continuous-reference V1/V2 adapters, we observed improved visual coherence, fewer clothing-color confusions, and stronger facial consistency in the tested cases. Since the versions change several design elements together, these observations concern the combined representation and do not identify the contribution of independent encoding, the MLP, or the temporal offsets individually. All compared variants were trained on the same LTX backbone, so the version comparison does not involve a change of training backbone.

The same testing revealed recurring limitations, summarized in \cref{tab:failures}. Simple clothing, uncomplicated scene references, and objects were easier to maintain in the tested examples. Visually similar characters and garments remained more difficult. For example, two same-gender characters wearing similarly shaped short jackets could exhibit appearance blending or clothing exchange, even when jacket colors differed. A gray-jacket character could become inconsistent after turning or after a cut away and back. Elaborate dresses with overlapping sleeve and skirt structures were another observed failure case. These are development observations; their frequencies were not logged.

\begin{table}[t]
\centering\small
\begin{tabularx}{\linewidth}{@{}p{.27\linewidth}X@{}}\toprule
Condition & Observed behavior\\\midrule
Simple clothing, props, scenes & Reference appearance was often maintained in tested examples.\\
Similar character appearance or clothing & Subject blending and clothing exchange could occur.\\
Turning or returning after a cut & Appearance could become inconsistent after the viewpoint change.\\
Complex overlapping garment structures & Sleeve and skirt structures could merge or be misinterpreted.\\\bottomrule
\end{tabularx}
\caption{Qualitative development observations. No event counts or failure probabilities were retained, so the rows do not imply measured rates.}
\label{tab:failures}
\end{table}

The failures show why a slot tag should not be interpreted as a hard identity constraint. It helps identify the source of conditioning tokens, but the transformer still has to associate that source with the correct generated region and preserve fine structure during motion. Source indexing and successful visual binding are distinct requirements.

\subsection{Supplementary audio-reference experiment}
\label{sec:audio}
We also extend MSR with audio-reference conditioning as a supplementary experiment. The additional stage freezes the visual parameters, including the visual slot MLP, and trains audio-related adapters and an audio-slot module. This adds voice-reference conditioning to the visual multi-subject system without updating its visual weights. In exploratory tests, explicit text assignments associated two speakers with two reference voices. The tests also revealed voice-characteristic mixing, inherited reference noise, and artifacts during emotional speech. No obvious visual change was observed in comparisons using the same images, prompts, and inference backbone, although random seeds were not matched. These observations are qualitative and do not establish speaker-similarity scores or strict visual equivalence. Appendix~\ref{app:audio} details the audio conditioning, training setup, and observed limitations.

\FloatBarrier

%% file: figures/case_03.tex
\begingroup\setlength{\tabcolsep}{2pt}\small\begin{tabular}{@{}ccc@{}}
Reference 1 & Reference 2 & Reference 3\\
\includegraphics[width=.327\linewidth]{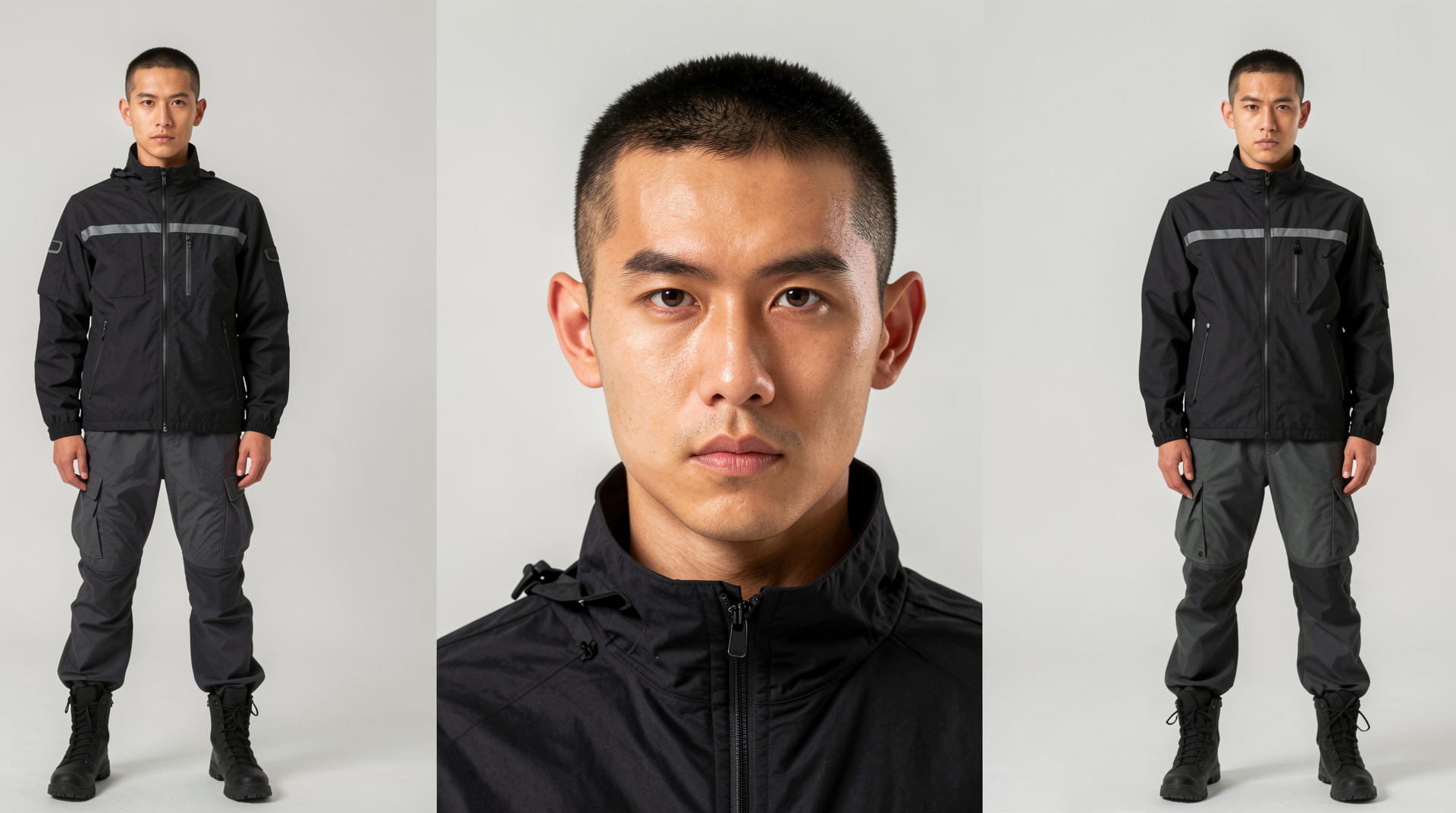} & \includegraphics[width=.327\linewidth]{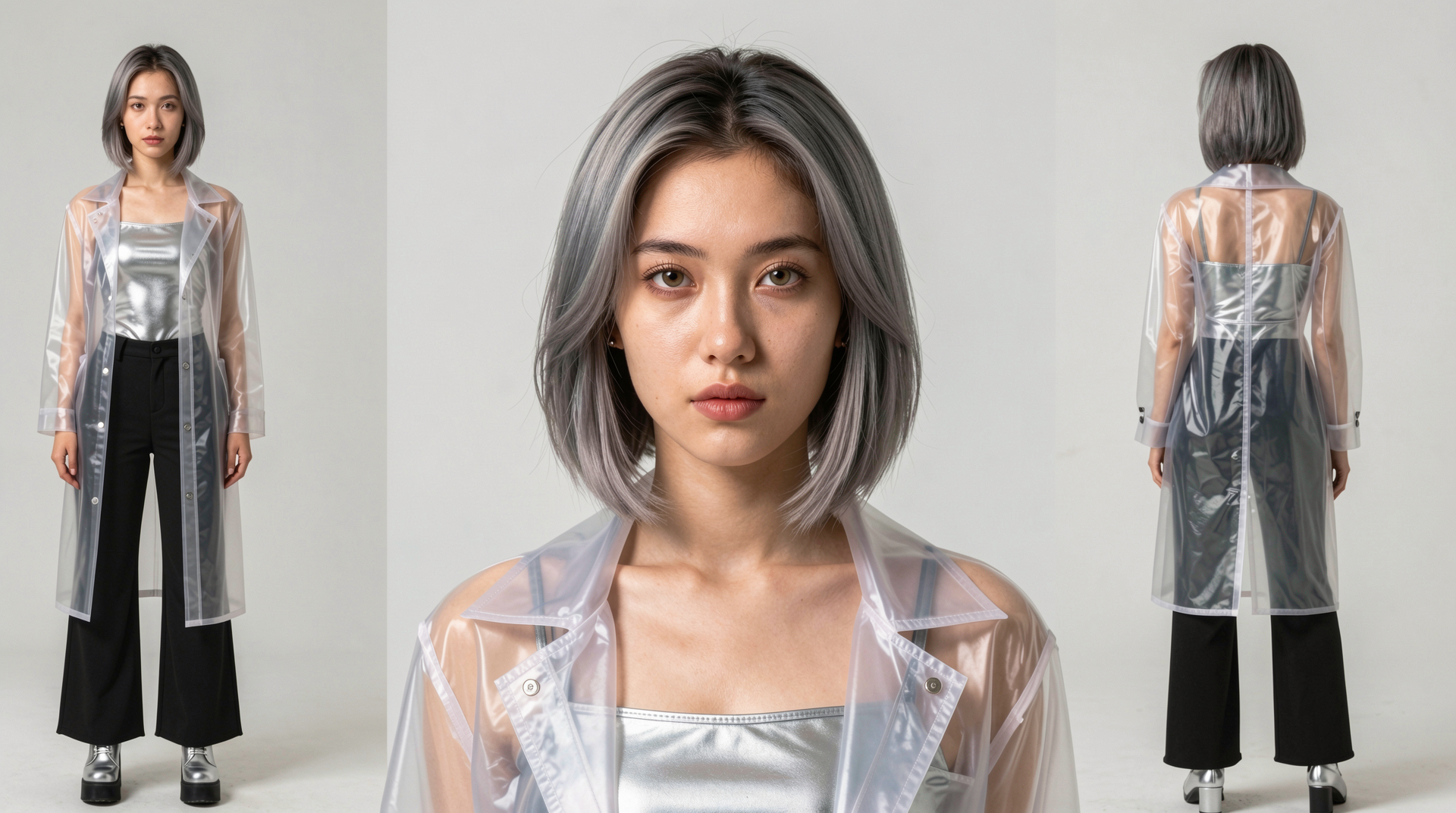} & \includegraphics[width=.327\linewidth]{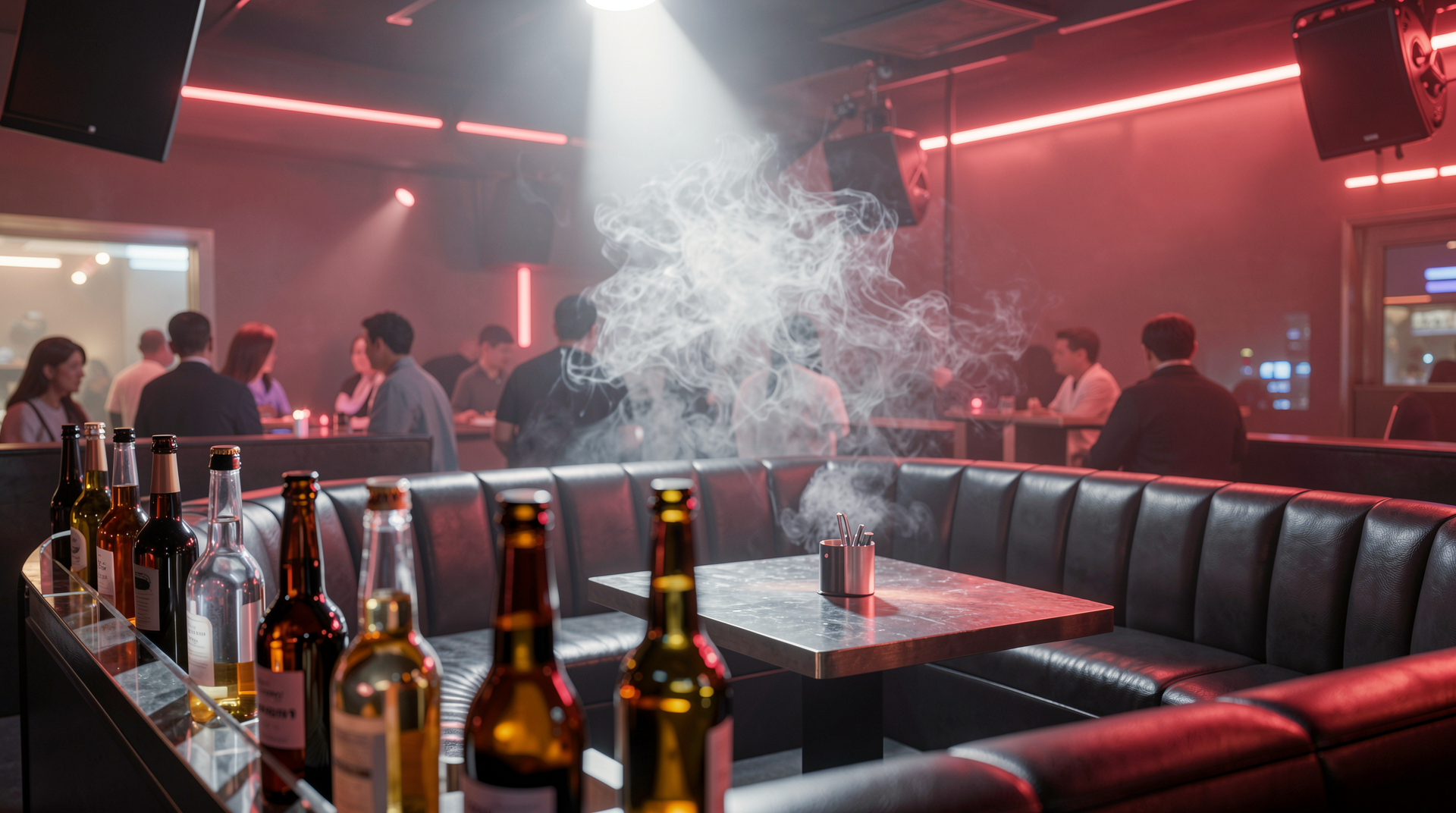}\\[3pt]
Generated: 1.83 s & Generated: 6.04 s & Generated: 10.25 s\\
\includegraphics[width=.327\linewidth]{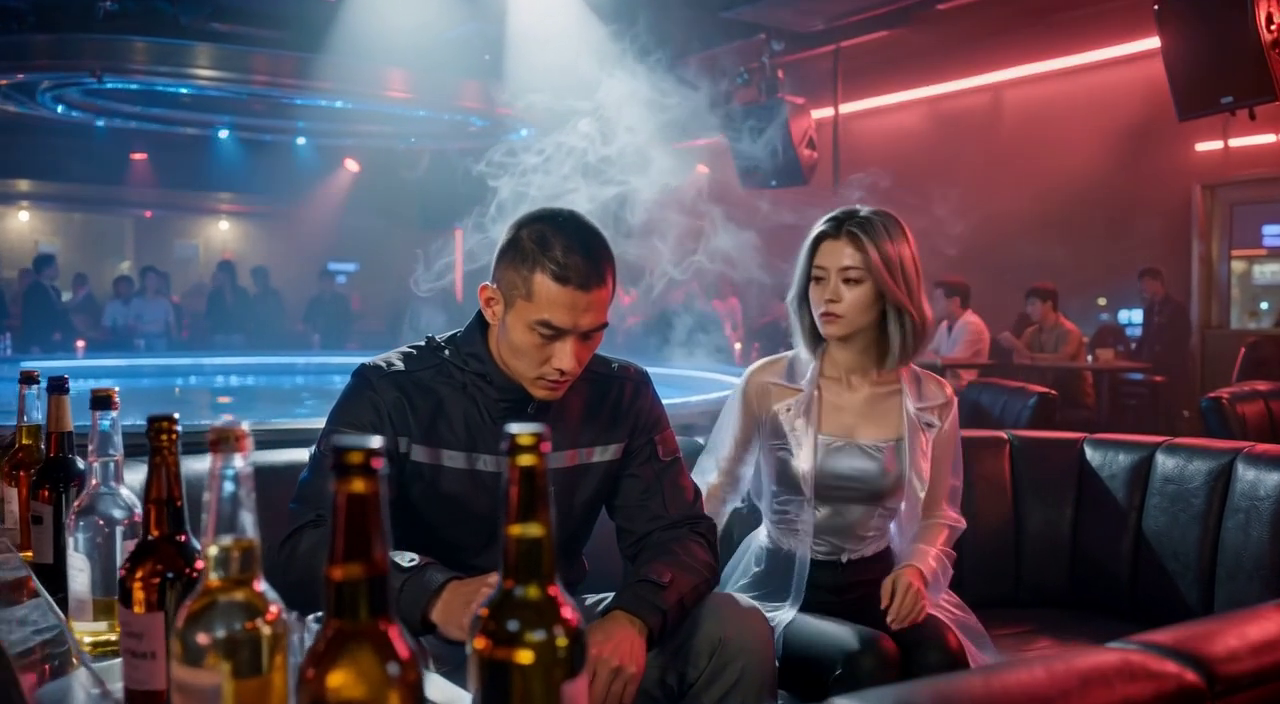} & \includegraphics[width=.327\linewidth]{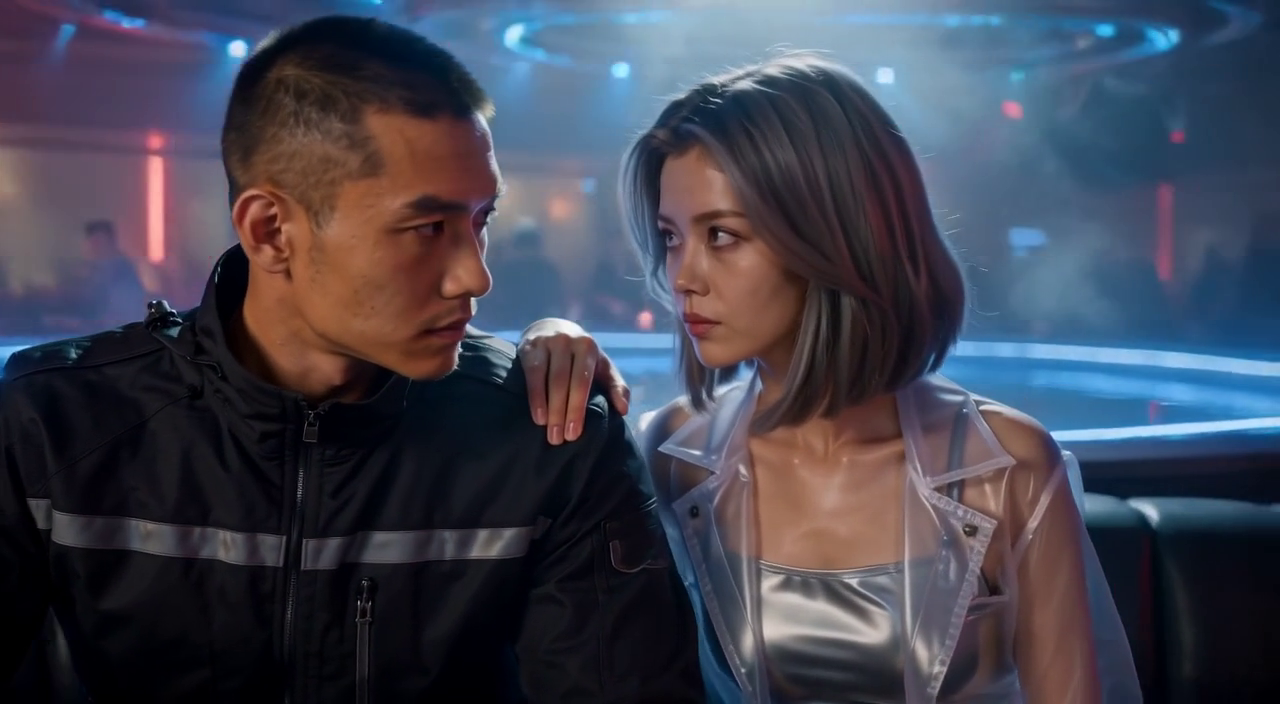} & \includegraphics[width=.327\linewidth]{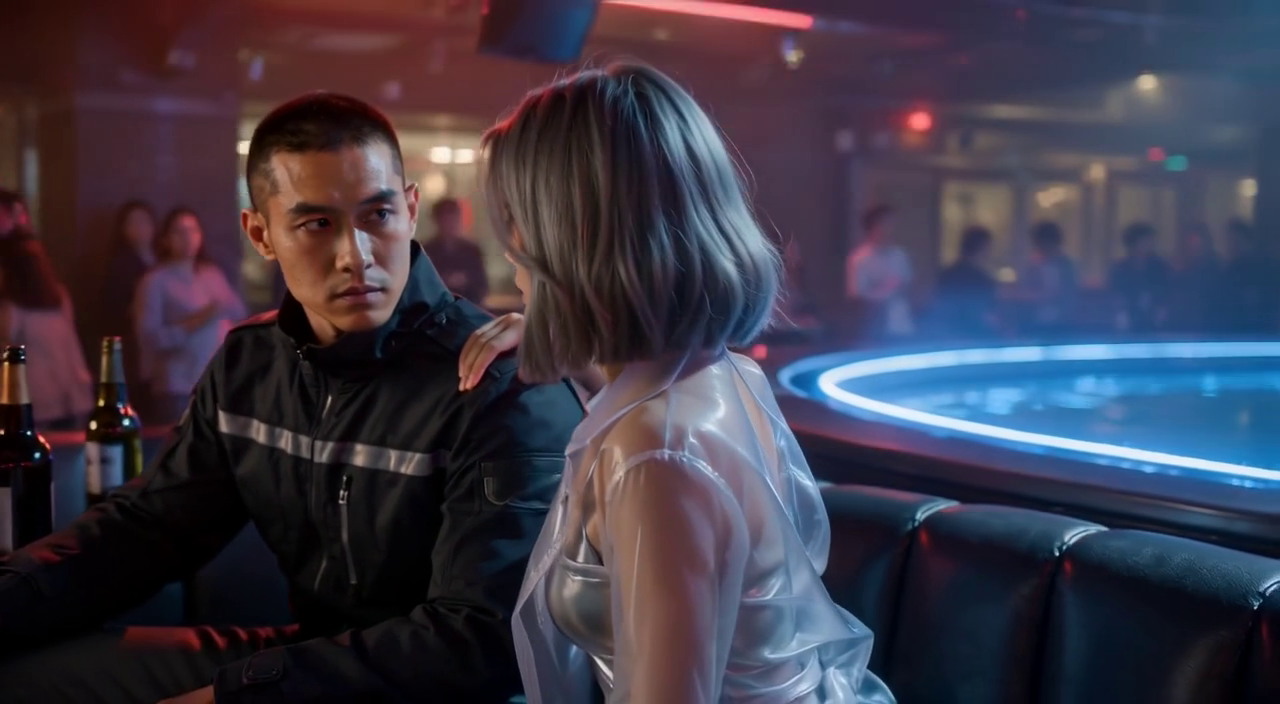}\\
\end{tabular}\endgroup

%% file: figures/case_06.tex
\begingroup\setlength{\tabcolsep}{2pt}\small\begin{tabular}{@{}ccc@{}}
Reference 1 & Reference 2 & Reference 3\\
\includegraphics[width=.327\linewidth]{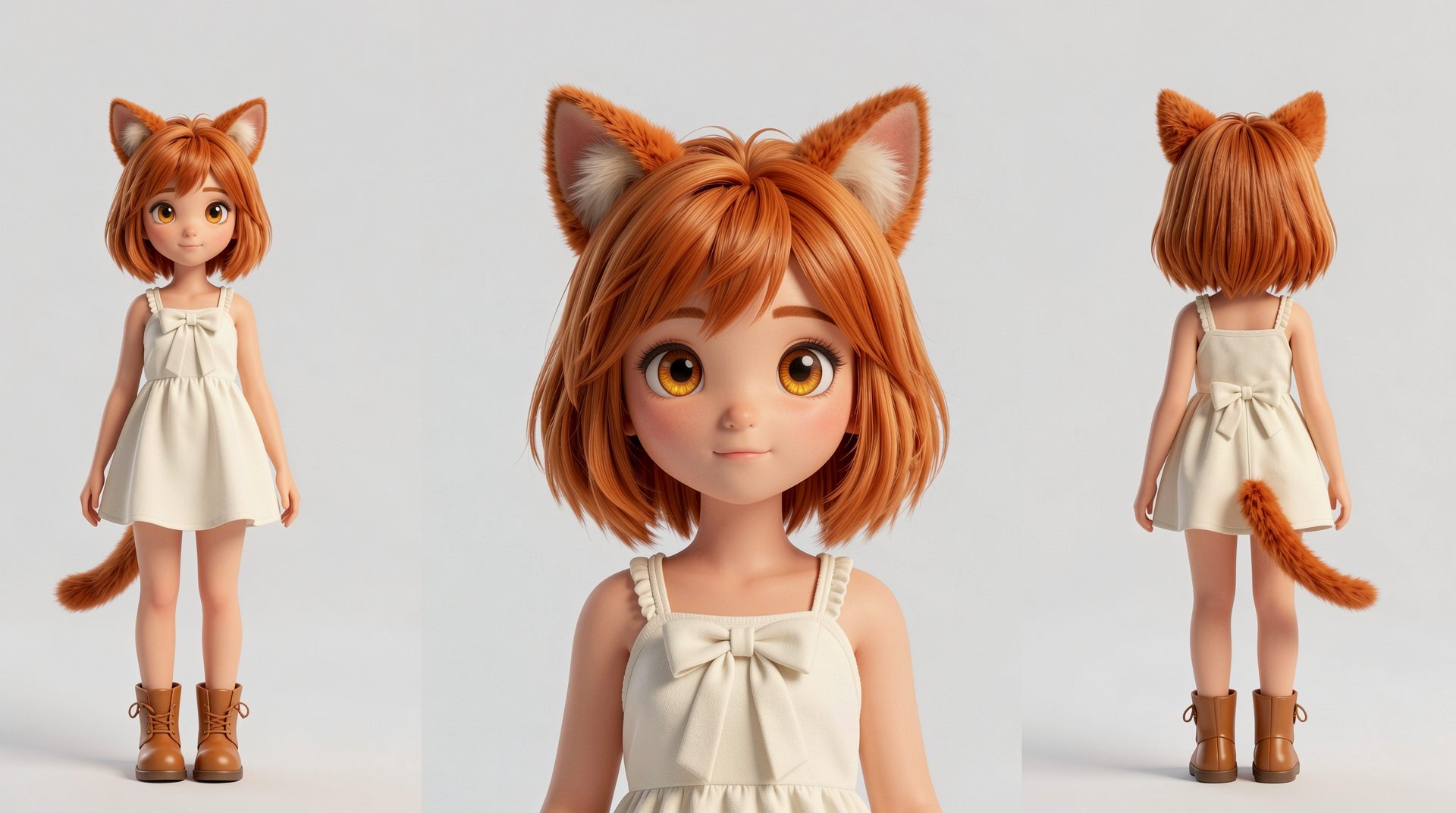} & \includegraphics[width=.327\linewidth]{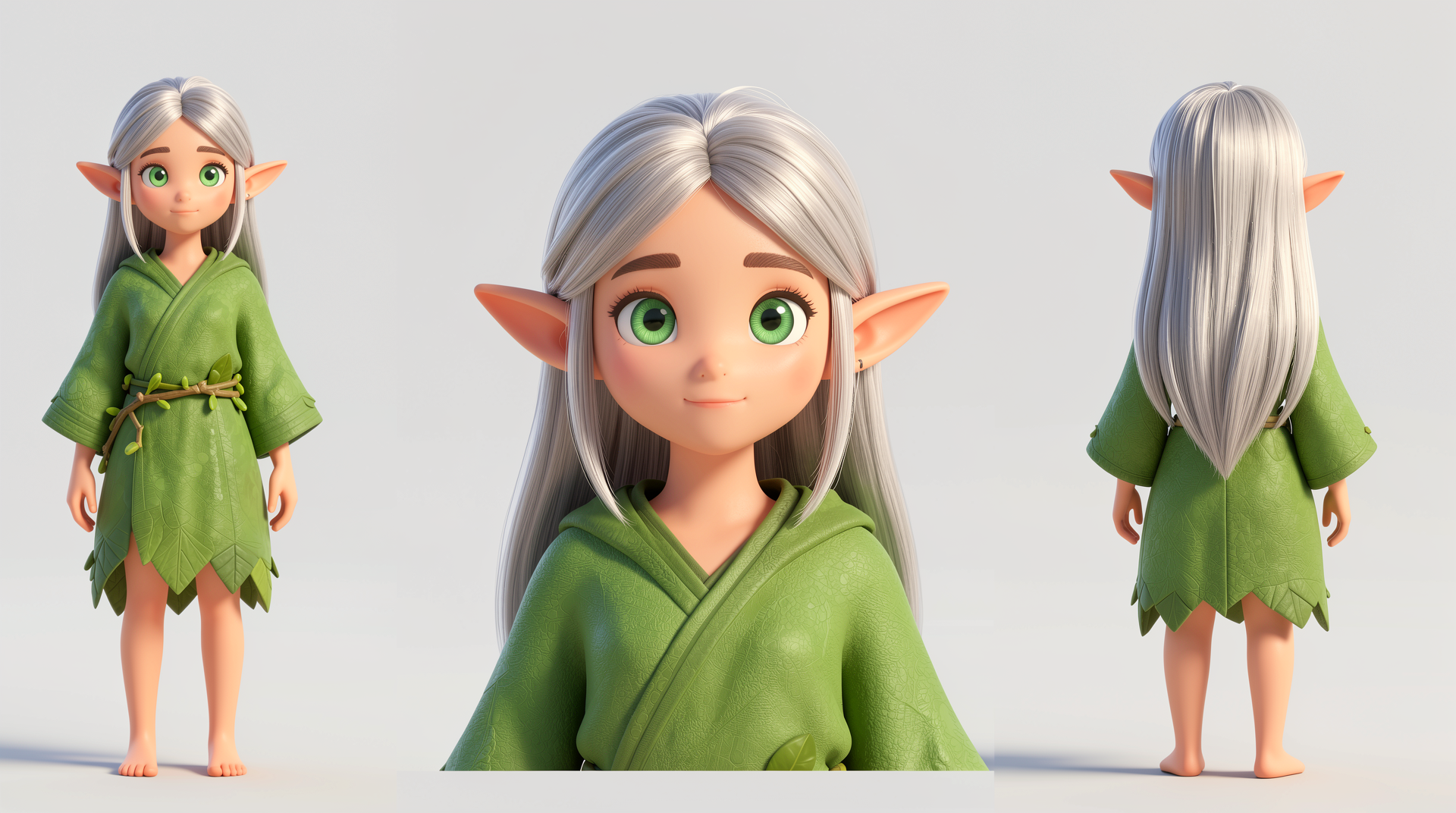} & \includegraphics[width=.327\linewidth]{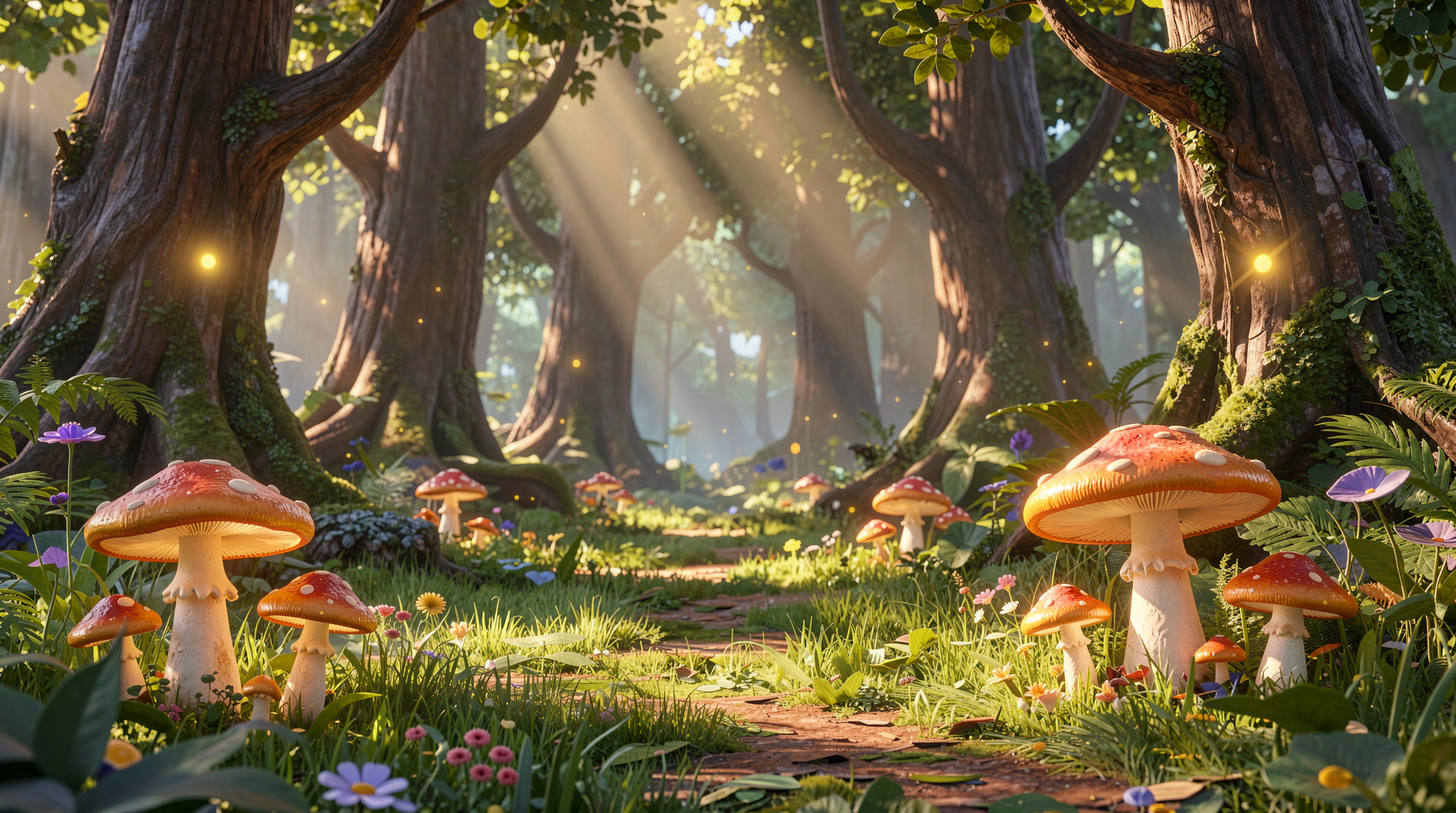}\\[3pt]
Generated: 1.83 s & Generated: 6.04 s & Generated: 10.25 s\\
\includegraphics[width=.327\linewidth]{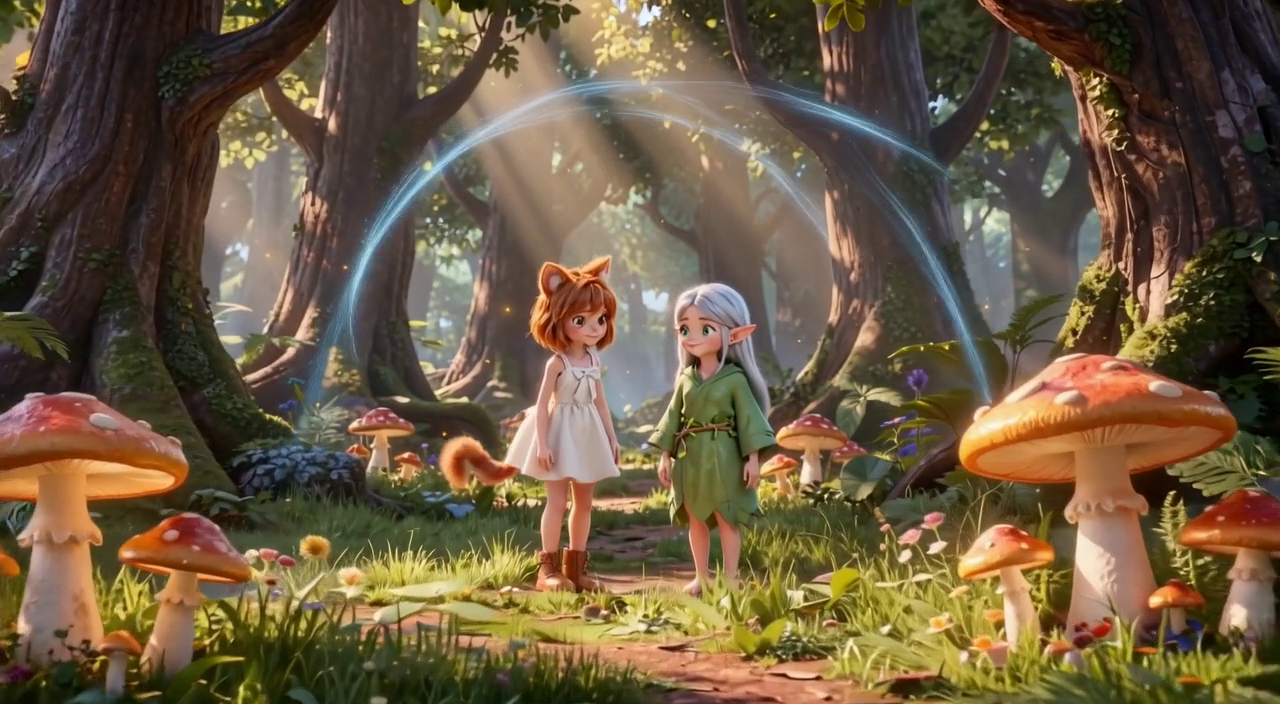} & \includegraphics[width=.327\linewidth]{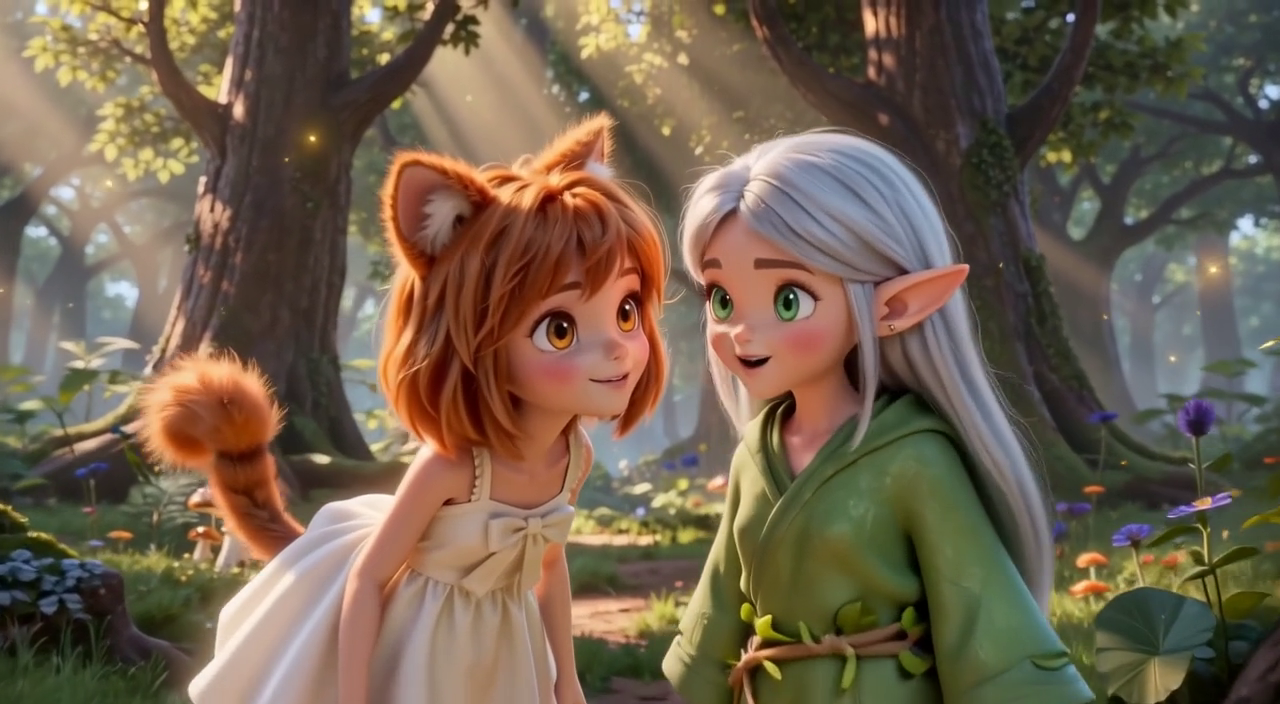} & \includegraphics[width=.327\linewidth]{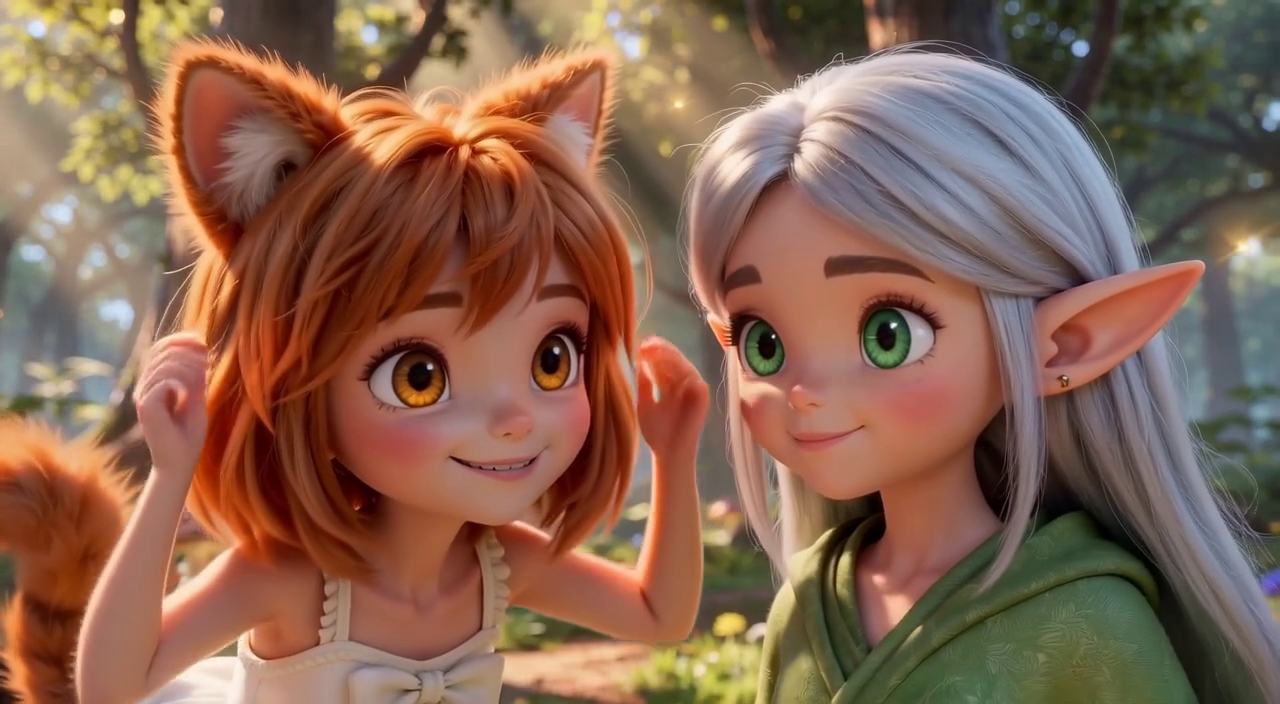}\\
\end{tabular}\endgroup

%% file: sections/discussion.tex
\section{Discussion and Limitations}
\paragraph{What the representation provides.}
Independent encoding makes reference boundaries explicit before transformer processing. The numeric MLP contributes a shared source signal to each group, and the temporal offset changes its positional context. The cues operate at feature and positional levels. They identify conditioning sources but do not prescribe the spatial regions occupied by generated subjects; assigning reference attributes to those regions remains a learned part of generation.

\paragraph{Evaluation and reproducibility.}
The present qualitative study lacks a retained seed-matched benchmark, isolated retraining ablations, a formal user study, and a quantitative comparison with other systems. The development corpus is not presented as a benchmark with a documented identity-disjoint test split. The released intermediate checkpoint also lacks a recoverable exact training-step index. These limitations restrict both statistical conclusions and exact reproduction of its training history. A future evaluation should separately measure appearance fidelity, role assignment, temporal behavior, and scene composition on a fixed set of inputs, while preserving outputs and run configurations.

\paragraph{Scope of reference counts and prompting.}
The public workflow accepts up to five reference images; generalization to larger numbers has not been established. Indexing also does not eliminate the need to describe intended roles. In practice, prompts identify each image, establish the scene and actor relationships, and describe actions and camera changes in order. These are useful interface conventions rather than a claim that one exact prompt syntax is necessary. The model can still misinterpret a complex garment or confuse similar-looking subjects even when the references are explicitly named.

\paragraph{Responsible use.}
Reference-conditioned synthesis can be used to depict recognizable people or imitate distinctive creative assets. Users should have appropriate permission to use and distribute the supplied references and should identify generated material when presenting it as such. The model's ability to reproduce appearance does not establish consent, ownership, or factual authenticity. Audio-reference use additionally requires care with speaker consent. This paper makes no claim that the training assets are unrestricted for redistribution.

\section{Conclusion}
We presented MSR, a slot-aware conditioning scheme for multi-subject reference video generation. Independently encoded reference groups receive numeric slot embeddings and temporal-coordinate offsets, then provide clean context for target-only flow-matching training. Qualitative examples show the composition of distinct characters and referenced environments in realistic and stylized videos. Development observations suggest reduced reference confusion relative to an earlier continuous-reference design, with similar clothing, complex garments, and viewpoint changes remaining challenging. A supplementary audio-reference extension adds voice conditioning through audio-related training with frozen visual parameters. The released adapter and workflows provide an implementation for further evaluation of reference fidelity, role assignment, and temporal consistency.

\paragraph{Availability.}
The visual adapter and ComfyUI workflows are available at \href{https://huggingface.co/LiconStudio/LTX-2.5-Multiple-Subject-Reference}{the public MSR model repository}. The earlier adapters are available at \href{https://huggingface.co/LiconStudio/LTX-2.3-Multiple-Subject-Reference}{the earlier MSR model repository}. A hosted inference interface is available through \href{https://huggingface.co/spaces/hugging-apps/ltx25-multi-subject-reference}{the hosted MSR demonstration}. The adapters described here are based on LTX.

%% file: sections/appendix.tex
\section{Implementation and Training Details}
\label{app:settings}
\paragraph{Reference preprocessing.}
The archived image-preprocessing script specifies buckets of $1280\times704$, $704\times1280$, $704\times704$, and $1280\times1280$, each with 25 repeated image frames. It uses centered resizing/cropping, VAE tiling, and separate latent directories for numbered image slots. The target-video path preserves source frame rates and selects supported $8n+1$ temporal lengths. The released example workflow and hosted interface default to 33 repeated frames per reference, whereas the documented training preprocessing uses 25. The exact settings of each published demonstration were not retained in its paired asset files. These are preprocessing settings, not a claim that every generated example uses the same output resolution or duration. Some source images are multi-view sheets; a sheet occupies one slot and is not split into multiple identities by the method.

\paragraph{Temporal alignment.}
For the archived training implementation, a reference coordinate $\tau^{\rm ref}_{i,n}$ is formed from the causal VAE grid at the stored reference frame rate. Before applying \cref{eq:offset}, the implementation uses
\begin{equation}
 \bar\tau_{i,n}=\max\left\{\tau^{\rm ref}_{i,n}-\frac{S_i-1}{f_v},\,0\right\},
\end{equation}
when $S_i>1$, where $S_i$ is the positive rounded ratio between target and reference latent temporal-group counts after excluding the first causal group. For $S_i=1$, the nonnegative reference coordinates are unchanged. This operation applies to both interval endpoints. Generic inference helpers additionally account for temporal scaling while constructing the grid. Thus \cref{eq:offset} describes the shared slot translation applied to the pre-aligned grid, rather than asserting identical grid construction for every training and inference resize configuration.

\paragraph{Noise sampling.}
The retained visual-stage configuration uses a two-region sampler. Let $U\in[0,1]$ be drawn from the implementation's length-aware, stretched and clipped shifted-logit-normal sampler with a 10\% uniform mixture, standard deviation 1, and endpoint parameter $\varepsilon=0.001$. It maps the draw into a low- or high-noise region as
\begin{equation}
 \sigma=\begin{cases}
 0.45U, & \text{with probability }0.6,\\
 0.45+0.55U, & \text{with probability }0.4.
 \end{cases}
\end{equation}
The sampler changes the distribution of training noise levels; it does not introduce an additional per-token loss weight. We record it as an implementation setting and do not claim a separately measured gain from this choice.

\paragraph{Checkpoint and release correspondence.}
The public visual weight is an intermediate checkpoint of the documented visual stage. Its original local file was removed after release. The later retained checkpoint and resume configuration are useful training records, but they are not substitutes for the exact release history. The public weight contains five tensors for the slot module, including its fixed frequency buffer. The two earlier adapters do not contain this learned slot module. All versions were trained using the same LTX backbone.

\section{Public Examples and Prompt Organization}
\label{app:examples}
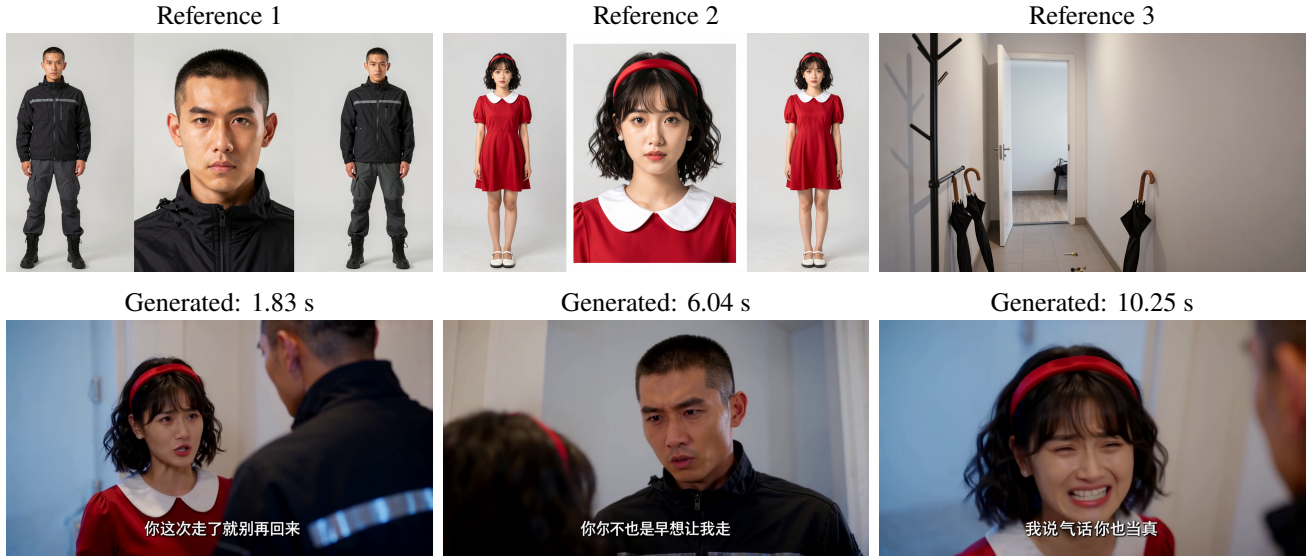
\begin{figure}[htbp]
\centering\input{figures/case_07.tex}
\caption{Additional public visual MSR case 07. Top: reference images. Bottom: three full frames from the published clip. Framing changes emphasize different characters, while several salient appearance cues remain visible. The subtitles are part of the supplied output. The figure does not evaluate the later audio-reference extension.}
\label{fig:cuts}
\end{figure}

Figures~\ref{fig:realistic}, \ref{fig:stylized}, and \ref{fig:cuts} use cases 03, 06, and 07 from the visual model repository's \code{validition\_V1} directory. The assets were retrieved at revision \code{9a053e6d63e54cd6970b1cda9419f1339457bd8c}. Reference images are shown as supplied; generated frames are sampled without cropping by selecting the first source frame at or after 15\%, 50\%, and 85\% of each 12.04-second clip. The nominal sampling times are 1.806, 6.020, and 10.234 seconds; the selected source frames occur at 1.833333, 6.041667, and 10.250000 seconds. The original videos and prompts provide context beyond the sampled frames.

\Cref{fig:cuts} shows another public case with two character references and an interior scene. The sampled frames alternate the visible focal character. The clothing colors and the red headband remain recognizable visual cues. Subtitles are already present in the source video and are retained in the figure; we do not use their appearance as an evaluation of speech, transcription, or audio-reference conditioning.

A useful prompt identifies each image and its role, establishes the setting and initial actor positions, and describes actions, object ownership, and camera changes in sequence. Image numbers refer to input sources rather than identity-recognition outputs. We used this organization during development, but did not establish that every clause is necessary through independently retained deletion experiments.

\section{Supplementary Audio-Reference Extension}
\label{app:audio}
The subsequent audio-reference stage is a separate extension of the visual system. It freezes visual parameters, including the visual slot MLP, and trains audio self-attention, audio text cross-attention, audio feed-forward adapters, video-to-audio attention adapters, and an audio-slot module. Video is conditioning-only during this training stage; the audio stream is generated and supplies the optimization loss. The recorded audio-stage training mixture comprises approximately 64.56\% from \code{short\_drama\_audio} and 35.44\% from \code{audio\_reference\_v2}. This distribution is reported separately from the visual-stage mixture.

Audio references use a separate temporal arrangement. The configuration allocates five-second audio slots with an end margin of 0.04 seconds and truncates overlong references. These audio windows should not be confused with the small visual offsets in \cref{eq:offset}. Joint appearance and voice personalization is also studied by ID-LoRA~\citep{IDLoRA2026}; our main contribution and qualitative figures concern the visual reference scheme.

In exploratory tests, two speakers were associated with two reference voices using explicit text assignments. With a missing voice reference, the model generated audio from the remaining context. Observed issues included mixtures of reference and model-generated voice characteristics, inherited reference noise, and synthetic or intermittent electronic-sounding artifacts during emotional speech. Cross-language observations were not sufficiently retained to support a specific conclusion, and generalization to unseen voices was not evaluated in the recorded tests. We therefore do not report a speaker-similarity or lip-synchronization score.

Comparisons with the visual version used the same images, prompts, and inference backbone but did not fix identical random seeds. No obvious visual change was observed in those tests. Freezing visual weights does not mathematically imply identical generated videos, because audiovisual inference can still couple the streams. The observation is consequently not a claim of strict visual equivalence or guaranteed absence of degradation.

%% file: figures/case_07.tex
\begingroup\setlength{\tabcolsep}{2pt}\small\begin{tabular}{@{}ccc@{}}
Reference 1 & Reference 2 & Reference 3\\
\includegraphics[width=.327\linewidth]{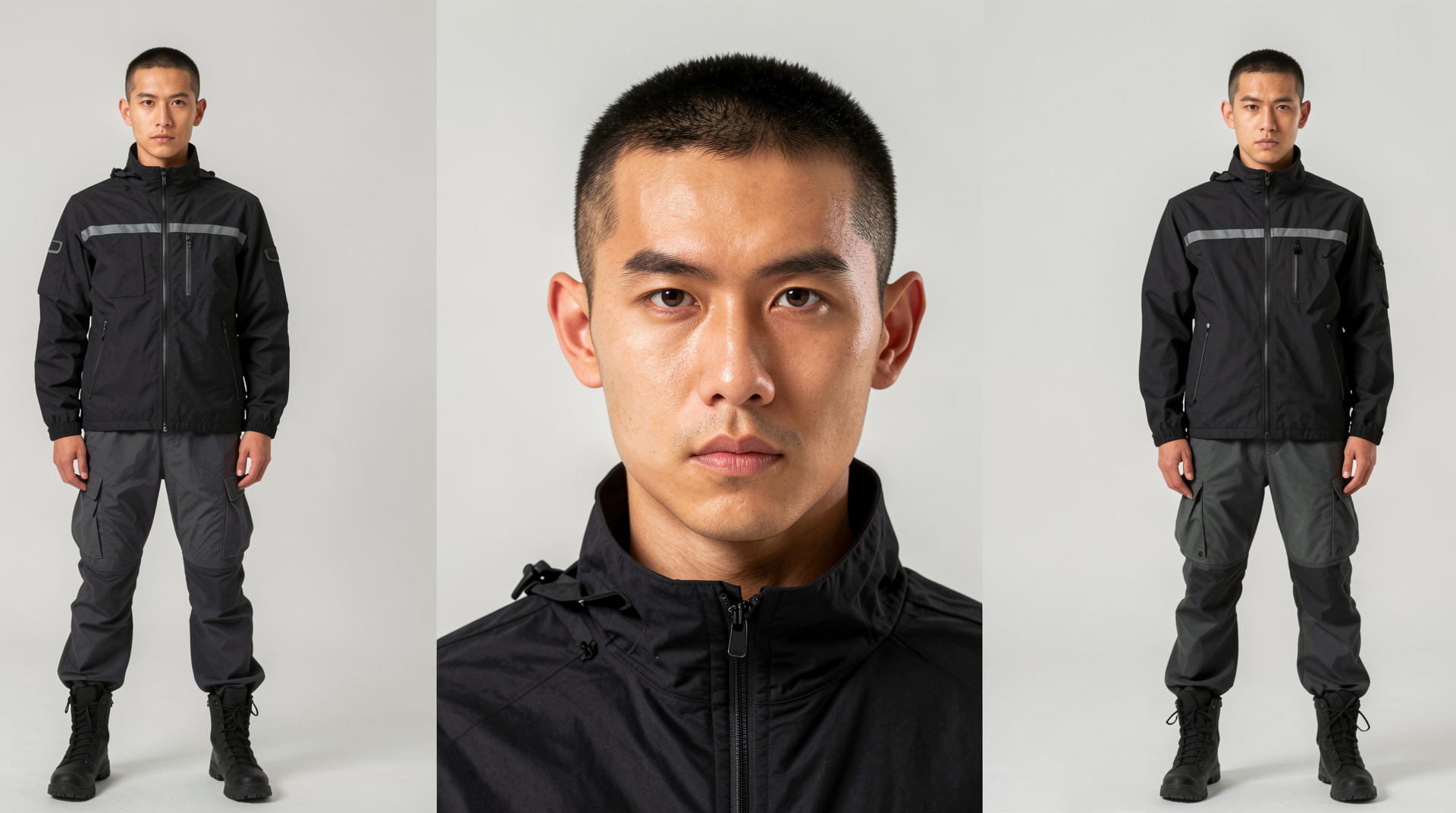} & \includegraphics[width=.327\linewidth]{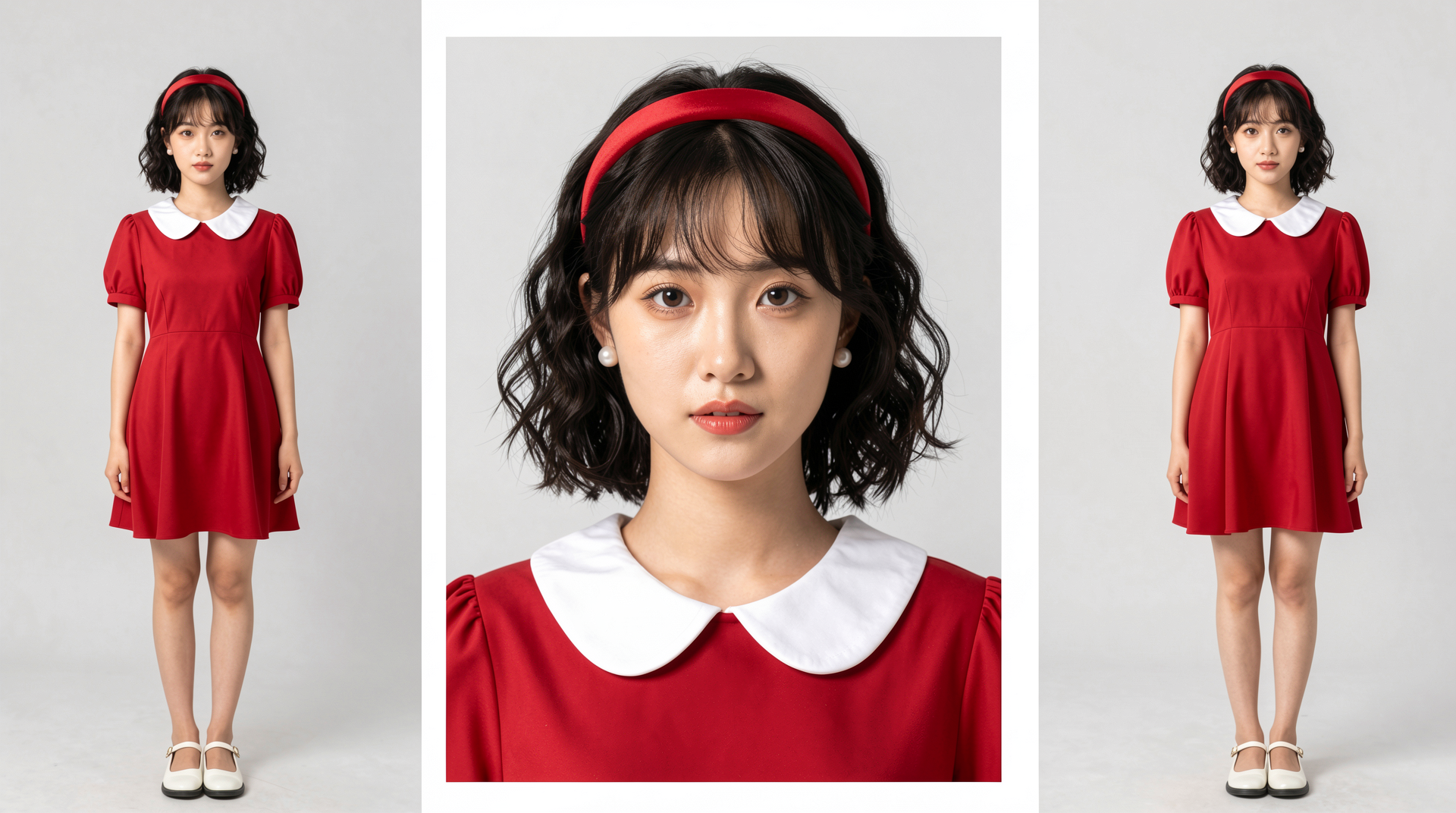} & \includegraphics[width=.327\linewidth]{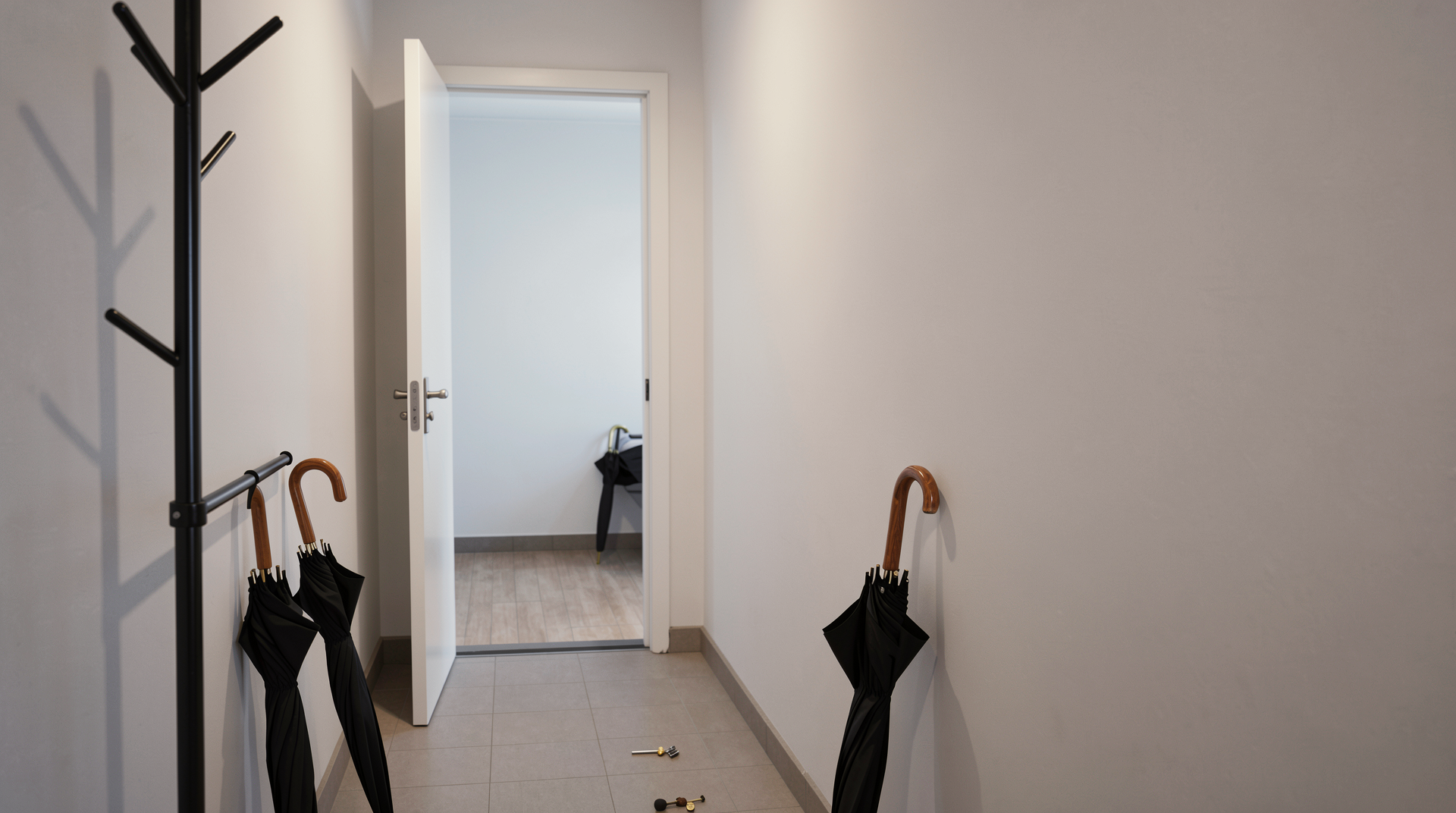}\\[3pt]
Generated: 1.83 s & Generated: 6.04 s & Generated: 10.25 s\\
\includegraphics[width=.327\linewidth]{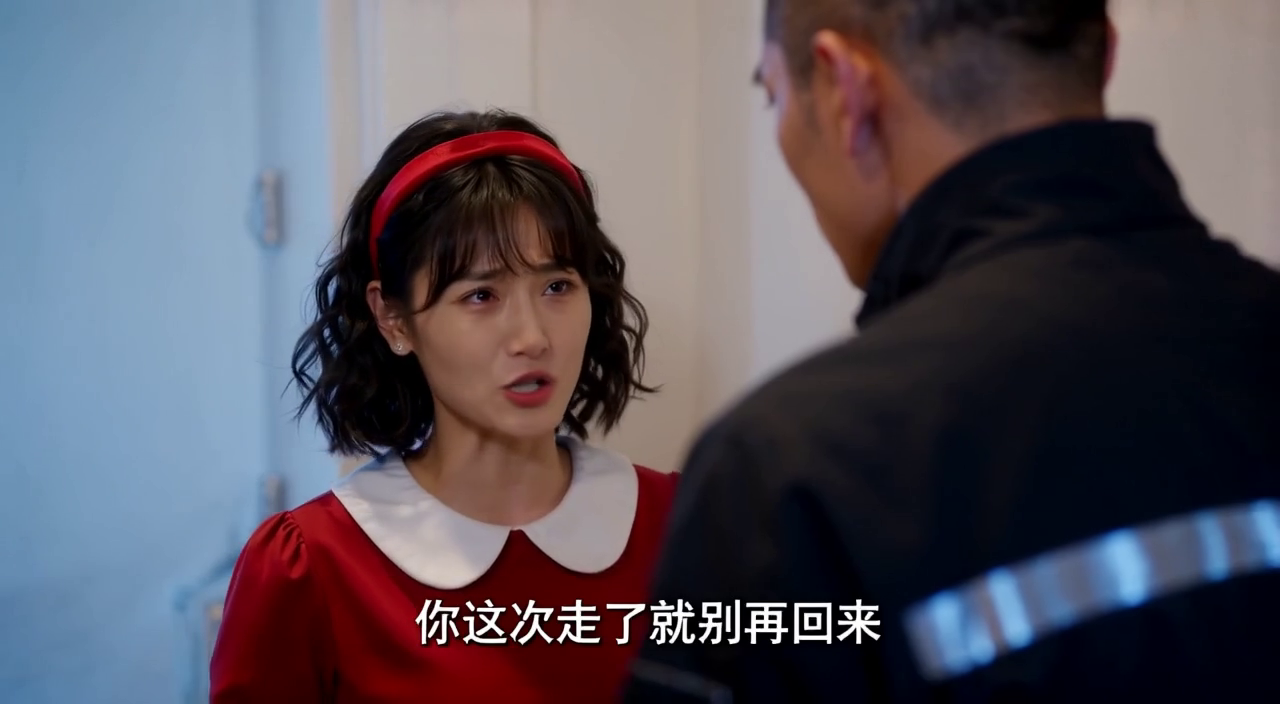} & \includegraphics[width=.327\linewidth]{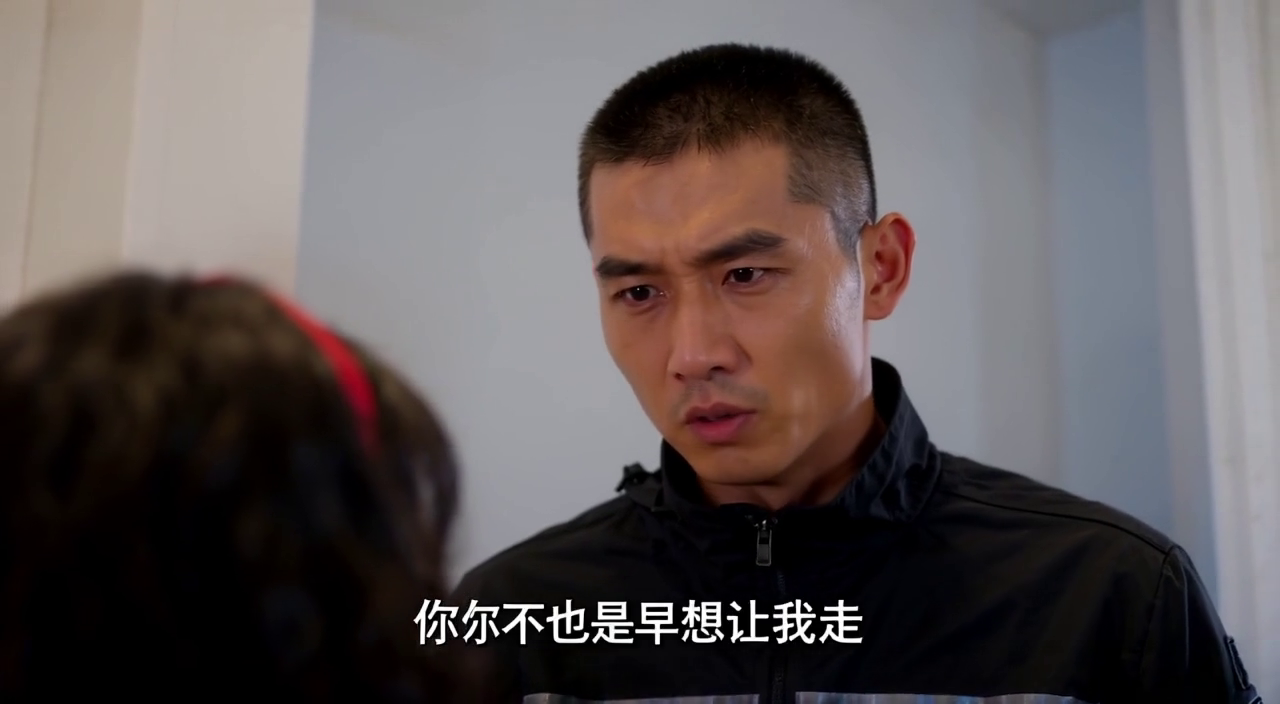} & \includegraphics[width=.327\linewidth]{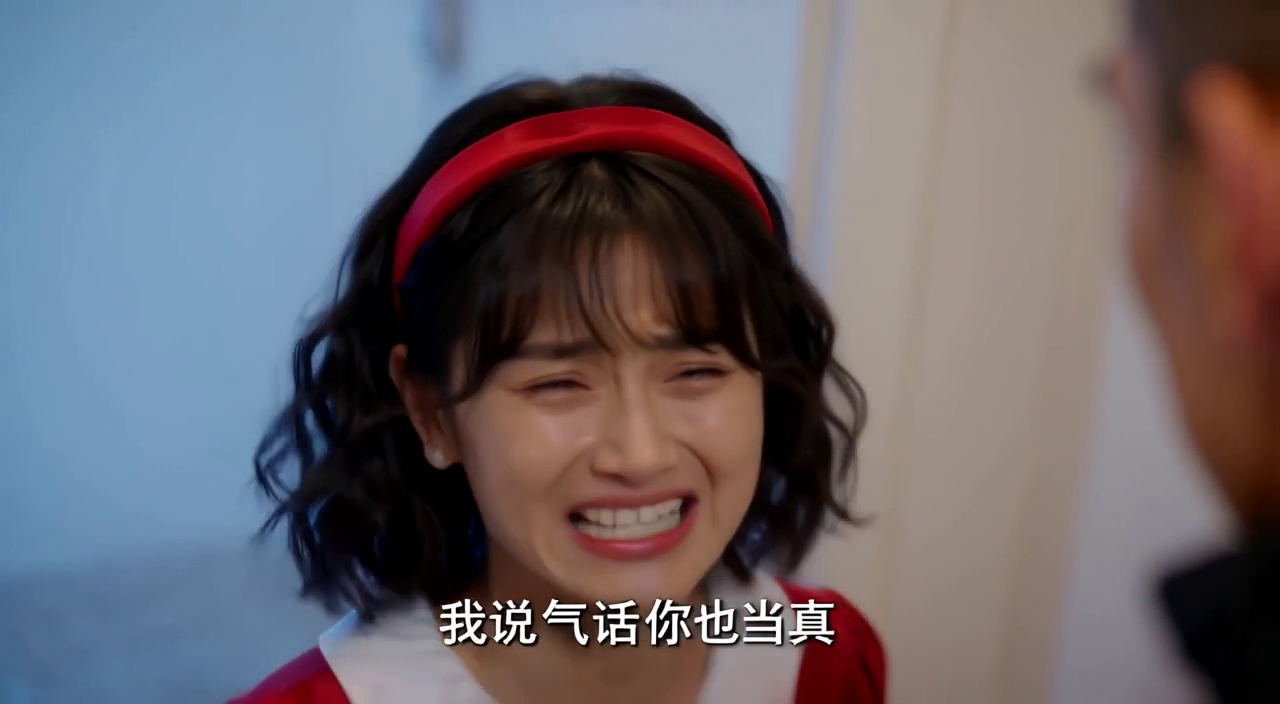}\\
\end{tabular}\endgroup

%% file: main.bbl
\begin{thebibliography}{9}
\providecommand{\natexlab}[1]{#1}
\providecommand{\url}[1]{\texttt{#1}}
\expandafter\ifx\csname urlstyle\endcsname\relax
  \providecommand{\doi}[1]{doi: #1}\else
  \providecommand{\doi}{doi: \begingroup \urlstyle{rm}\Url}\fi

\bibitem[Dahan et~al.(2026)Dahan, Yanuka, Kraicer, Wolf, and
  Giryes]{IDLoRA2026}
Aviad Dahan, Moran Yanuka, Noa Kraicer, Lior Wolf, and Raja Giryes.
\newblock {ID-LoRA}: Identity-driven audio-video personalization with
  in-context {LoRA}, 2026.
\newblock URL \url{https://arxiv.org/abs/2603.10256}.

\bibitem[Deng et~al.(2025)Deng, Yin, Guo, Wang, Fang, Yuan, Yang, Wang, Liu,
  Huang, and Ma]{MAGREF2025}
Yufan Deng, Yuanyang Yin, Xun Guo, Yizhi Wang, Jacob~Zhiyuan Fang, Shenghai
  Yuan, Yiding Yang, Angtian Wang, Bo~Liu, Haibin Huang, and Chongyang Ma.
\newblock {MAGREF}: Masked guidance for any-reference video generation with
  subject disentanglement, 2025.
\newblock URL \url{https://arxiv.org/abs/2505.23742}.

\bibitem[Fei et~al.(2025)Fei, Li, Qiu, Yu, and Fan]{Ingredients2025}
Zhengcong Fei, Debang Li, Di~Qiu, Changqian Yu, and Mingyuan Fan.
\newblock Ingredients: Blending custom photos with video diffusion
  transformers, 2025.
\newblock URL \url{https://arxiv.org/abs/2501.01790}.

\bibitem[HaCohen et~al.(2026)HaCohen, Brazowski, Chiprut, Bitterman, Kvochko,
  Berkowitz, Shalem, Lifschitz, Moshe, Porat, Richardson, Shiran, Chachy,
  Chetboun, Finkelson, Kupchick, Zabari, Guetta, Kotler, Bibi, Gordon, Panet,
  Benita, Armon, Kulikov, Inger, Shiftan, Melumian, and Farbman]{LTX22026}
Yoav HaCohen, Benny Brazowski, Nisan Chiprut, Yaki Bitterman, Andrew Kvochko,
  Avishai Berkowitz, Daniel Shalem, Daphna Lifschitz, Dudu Moshe, Eitan Porat,
  Eitan Richardson, Guy Shiran, Itay Chachy, Jonathan Chetboun, Michael
  Finkelson, Michael Kupchick, Nir Zabari, Nitzan Guetta, Noa Kotler, Ofir
  Bibi, Ori Gordon, Poriya Panet, Roi Benita, Shahar Armon, Victor Kulikov,
  Yaron Inger, Yonatan Shiftan, Zeev Melumian, and Zeev Farbman.
\newblock {LTX-2}: Efficient joint audio-visual foundation model, 2026.
\newblock URL \url{https://arxiv.org/abs/2601.03233}.

\bibitem[Hu et~al.(2021)Hu, Shen, Wallis, Allen-Zhu, Li, Wang, Wang, and
  Chen]{LoRA2021}
Edward~J. Hu, Yelong Shen, Phillip Wallis, Zeyuan Allen-Zhu, Yuanzhi Li, Shean
  Wang, Lu~Wang, and Weizhu Chen.
\newblock {LoRA}: Low-rank adaptation of large language models, 2021.
\newblock URL \url{https://arxiv.org/abs/2106.09685}.

\bibitem[Li et~al.(2025)Li, Qian, Su, Diao, Xia, Liu, Yang, Zhang, and
  Yuan]{BindWeave2025}
Zhaoyang Li, Dongjun Qian, Kai Su, Qishuai Diao, Xiangyang Xia, Chang Liu,
  Wenfei Yang, Tianzhu Zhang, and Zehuan Yuan.
\newblock {BindWeave}: Subject-consistent video generation via cross-modal
  integration, 2025.
\newblock URL \url{https://arxiv.org/abs/2510.00438}.

\bibitem[Lipman et~al.(2022)Lipman, Chen, Ben-Hamu, Nickel, and
  Le]{FlowMatching2022}
Yaron Lipman, Ricky T.~Q. Chen, Heli Ben-Hamu, Maximilian Nickel, and Matt Le.
\newblock Flow matching for generative modeling, 2022.
\newblock URL \url{https://arxiv.org/abs/2210.02747}.

\bibitem[Su et~al.(2021)Su, Lu, Pan, Murtadha, Wen, and Liu]{RoFormer2021}
Jianlin Su, Yu~Lu, Shengfeng Pan, Ahmed Murtadha, Bo~Wen, and Yunfeng Liu.
\newblock {RoFormer}: Enhanced transformer with rotary position embedding,
  2021.
\newblock URL \url{https://arxiv.org/abs/2104.09864}.

\bibitem[Zhou et~al.(2026)Zhou, Yu, Ma, Wang, Yu, Wang, Yang, Chen, Ou, Liu,
  Zhou, and Lu]{Aura2026}
Zixiang Zhou, Zhentao Yu, Yifeng Ma, Hongmei Wang, Wenqing Yu, Cong Wang, Zilin
  Yang, Rui Chen, Jiarong Ou, Yezhou Liu, Yuan Zhou, and Qinglin Lu.
\newblock {Aura}: Consistent multi-subject video generation via {VLM}-grounded
  semantic alignment, 2026.
\newblock URL \url{https://arxiv.org/abs/2607.04311}.

\end{thebibliography}
